\documentclass[final]{colt2024} 

\title[Harmonics of Learning]{Harmonics of Learning: \\ Universal Fourier Features Emerge in Invariant Networks}
\usepackage{times}

\usepackage{amssymb,amsfonts,amsmath}
\usepackage{nicefrac}       %
\usepackage{microtype}      %
\usepackage{xcolor}         
\usepackage{enumerate,enumitem,tikz,graphicx,mathrsfs,eucal,verbatim, bbm, derivative}
\usepackage{float}
\usepackage[makeroom]{cancel}
\usepackage{transparent, xfrac}
\usepackage[labelfont=bf]{caption}

\newtheorem{infthm}[theorem]{Informal Theorem}

\newcommand{\free}[1]{\langle #1 \rangle}
\newcommand{\irr}[1]{\textnormal{Irr}(#1)}

\coltauthor{%
 \Name{Giovanni Luca Marchetti} \Email{glma@kth.se}\\
 \addr KTH Royal Institute of Technology
 \AND
 \Name{Christopher J Hillar} \Email{hillarmath@gmail.com}\\
 \addr Redwood Center for Theoretical Neuroscience 
 \AND
 \Name{Danica Kragic} \Email{dani@kth.se}\\
 \addr KTH Royal Institute of Technology
  \AND
 \Name{Sophia Sanborn} \Email{sophias@science.xyz}\\
 \addr Science Corporation
}

\begin{document}

\maketitle

\begin{abstract}
In this work, we formally prove that, under certain conditions, if a neural network is invariant to a finite group then its weights recover the Fourier transform on that group. This provides a mathematical explanation for the emergence of Fourier features -- a ubiquitous phenomenon in both biological and artificial learning systems. The results hold even for non-commutative groups, in which case the Fourier transform encodes all the irreducible unitary group representations. Our findings have consequences for the problem of symmetry discovery. Specifically, we demonstrate that the algebraic structure of an unknown group can be recovered from the weights of a network that is at least approximately invariant within certain bounds. Overall, this work contributes to a foundation for an algebraic learning theory of invariant neural network representations. \end{abstract}

\begin{keywords}
Invariant neural networks, harmonic analysis, group representations 
\end{keywords}

\begin{figure}[H]
    \centering
    {\transparent{.9}
    \includegraphics[width=\textwidth]{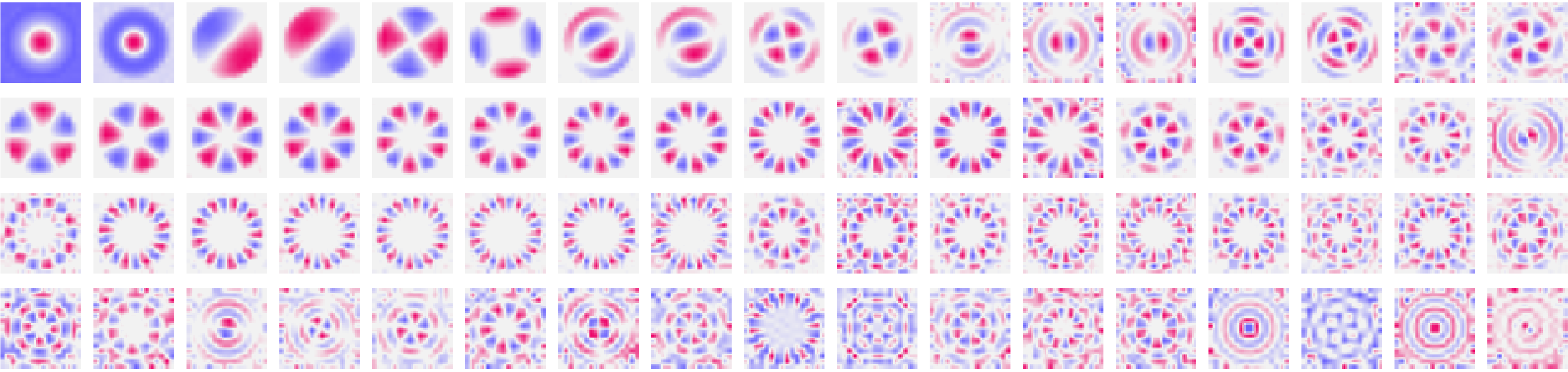}
    }
\caption{Weights learned by a neural network trained for invariance to planar rotations resemble circular harmonics. Data from \cite{sanborn2023bispectral}.}\label{fig:feature-examples}
\end{figure}

\section{Introduction and Related Work}

Artificial neural networks trained on natural data exhibit a striking phenomenon: regardless of exact initialization, dataset, or training objective, models trained on the same data domain frequently converge to similar learned representations \citep{li2015convergent}. For example, the early layer weights of diverse image models tend to converge to Gabor filters and color-contrast detectors \citep{olah2020overview}. Remarkably, many of these same features are observed in the visual cortex \citep{hubel1959receptive, hass2013v1, willeke2023deep}, suggesting a form of representational universality that transcends biological and artificial substrates. While such findings are empirically well-established \citep{rauker2023toward}, the field lacks theoretical explanations.

Spatially localized versions of canonical 2D Fourier basis functions, such as Gabor filters or wavelets, are perhaps the most frequently observed universal features. They commonly arise in the early layers of vision models -- trained with efficient coding \citep{olshausen1997sparse, bell1997independent}, classification \citep{olah2020overview}, temporal coherence \citep{hurri2003simple}, and next-step prediction \citep{fiquet2023polar} objectives --  as well as in the primary visual cortices of diverse mammals -- including cats \citep{hubel1962receptive}, monkeys \citep{hubel1968receptive}, and mice \citep{drager1975receptive}. Non-localized Fourier features have been observed in networks trained to solve tasks that permit cyclic wraparound -- for example, modular arithmetic \citep{nanda2023progress}, more general group compositions \citep{chughtai2023toy}, or invariance to the group of cyclic translations \citep{sanborn2023bispectral}. In the domain of spatial navigation, the so-called \textit{grid cells} of the entorhinal cortex \citep{moser2008place} display periodic firing patterns at different spatial frequencies as they build a map of space. Their response properties are naturally modeled with the harmonics of the twisted torus \citep{guanella2007model, orchard2013does, gardner2022toroidal}. Similar features also emerge in artificial neural networks trained to solve spatial navigation tasks \citep{banino2018vector, cueva2018emergence, sorscher2019unified, dorrell2022actionable, sorscher2023unified, schaeffer2023self}. The ubiquity of these features across diverse learning systems is both striking and unexplained.

In this work, we provide a mathematical explanation for the  emergence of Fourier features in learning systems such as neural networks. We argue that the mechanism responsible for this emergence is the downstream \textit{invariance} of the learner to the action of a \textit{group of symmetries} (e.g., planar translations or rotations). Since natural data typically possess symmetries, invariance is a fundamental bias that is injected both implicitly and sometimes explicitly into learning systems \citep{bronstein2021geometric, cohenc16, pmlr-v80-kondor18a}. Motivated by this, we derive theoretical guarantees for the presence of Fourier features in invariant learners that apply to a broad class of machine learning models. 

Our results rely on the inextricable link between harmonic analysis and group theory \citep{folland2016course}. The standard discrete Fourier transform is a special case of more general Fourier transforms on groups, which can be defined by replacing the standard basis of harmonics by irreducible unitary group representations. The latter are equivalent to the familiar definition for cyclic or, more generally, commutative groups, but are more involved for non-commutative ones. In order to accommodate both scenarios, we develop a general theory that applies, in principle, to arbitrary finite groups. 

This work represents an attempt to provide mathematical grounding for a general algebraic theory of representation learning, while addressing the \textit{universality hypothesis} for neural networks \citep{olah2020overview, moschella2022relative}. A suite of earlier theoretical works \citep{isely2010deciphering, hillar2015can, garfinkle2019uniqueness} established such universality for sparse coding models \citep{olshausen1997sparse}, deriving the conditions under which a network will recover the original bases that generate data through sparse linear combinations. In this case, the statistics of the data determine the uniqueness of the representation. Our findings, on the other hand, are purely \textit{algebraic}, since they rely exclusively on the invariance properties of the learner. Given the centrality of invariance to many machine learning tasks, our theory encompasses a broad class of scenarios and neural network architectures, while providing a new perspective on classical neuroscience \citep{pitts1947we, hubel1962receptive}. As such, it sets a foundation for a learning theory of representations in artificial and biological neural systems, grounded in the mathematics of symmetry.

\subsection{Overview of Results}\label{sec:overview}
In this section, we provide a non-technical overview of the theoretical results presented in this work. Our main result can be summarized as follows.  
\begin{infthm}[Theorem \ref{prop:mainnew} and Corollary \ref{mainpropnc}]\label{thm:mainintui}
If $\varphi(W,x)$ is a parametric function of a certain kind that is invariant in the input variable $x$ to the action of a finite group $G$, then each component of its weights $W$ coincides with a harmonic of $G$ up to a linear transformation. In particular, when the weights are orthonormal, $W$ coincides with the Fourier transform of $G$ up to linear transformations. 
\end{infthm}
Here, the term ``harmonic" refers to an irreducible unitary representation of $G$. In particular, one-dimensional unitary representations correspond to homomorphisms with the unit circle $\textnormal{U}(1) \subseteq \mathbb{C}$, which is reminiscent of the classical definition via the imaginary exponential. However, \emph{non-commutative} groups can have higher-dimensional irreducible representations, intuitively meaning that harmonics are valued in unitary matrices. In this case, the components of $W$ can be interpreted as \emph{capsules} in the sense of \cite{hinton2018matrix}; i.e., neural units processing matrix-valued signals. 

Since harmonic analysis is naturally formalized over the complex numbers, we consider models with complex weights $W$, which fits into the broader program of complex-valued machine learning \citep{bassey2021survey, trabelsi2017deep, lowe2022complex}. We show that the hypothesis on $\varphi$ in Theorem \ref{thm:mainintui} is satisfied by several machine learning models from the literature.  In particular, the theorem applies to the recently-introduced (Bi)Spectral Networks \citep{sanborn2023bispectral}, to single fully-connected layers of McCulloch-Pitts neurons, and, to an extent, to traditional deep networks. As an additional contribution, we generalize Spectral Networks to non-commutative groups.     

The group-theoretical Fourier transform encodes the entire group structure of $G$. Therefore, as a consequence of Theorem \ref{thm:mainintui}, the multiplication table of $G$ can be recovered from the weights $W$ of an invariant parametric function $\varphi$ -- a fact empirically demonstrated by \cite{sanborn2023bispectral}. This  addresses the question of \emph{symmetry discovery} -- an established machine learning problem aiming to recover the unknown group of symmetries of data with minimal supervision and prior knowledge \citep{rao1998learning, sohl2010unsupervised, desai2022symmetry}. Since the multiplication table is a discrete object, it is expected that the invariance constraint on $\varphi$ can be loosened while still recovering the group correctly. To this end, we prove the following. 

\begin{infthm}[Theorem \ref{prop:mainrelax}]\label{thm:boundsintui}
If $\varphi(W,x)$ is ``almost invariant" to $G$ according to certain functional bounds and the weights are ``almost orthonormal", then the multiplication table of $G$ can be recovered from $W$. 
\end{infthm}

Lastly, we implement a model satisfying the requirements of our theory and demonstrate its symmetry discovery capabilities. To this end, we train it via contrastive learning on an objective encouraging invariance and extract the multiplication table of $G$ from its weights. Our Python implementation is available at a public repository\footnote{\url{https://github.com/sophiaas/spectral-universality}}.       

\section{Mathematical Background}
\label{sec:background}

We begin by introducing the fundamental concepts from harmonic analysis and group theory used in this paper. For a complete treatment, we refer the reader to  \cite{steinberg2012representation}. 

\subsection{Groups and Actions}
\label{appendix:math}

A \textit{group} is an algebraic object whose elements represent abstract symmetries, which can be composed and inverted.

\begin{definition}\label{groupdef}
A group is a set $G$ equipped with a \emph{multiplication map} $G \times G \rightarrow G$ denoted by $(g,h) \mapsto gh$, an \emph{inversion map} $G \rightarrow G$ denoted by $g \mapsto g^{-1}$, and a distinguished \emph{identity element} $1 \in G$ such that for all $g, h, k \in G$:
\begin{center}
\begin{tabular}{ccc}
\emph{Associativity} &   \emph{Inversion} & \emph{Identity}  \\
 $g(hk) = (gh)k$ & \hspace{.2cm} $g^{-1}g = g g^{-1} = 1$  \hspace{.2cm} & $g1 = 1g = g$    
\end{tabular}
\end{center}
A map $\rho: \ G \rightarrow G'$ between groups is called a \emph{homomorphism} if $\rho(gh) = \rho(g) \rho(h)$ for all $g,h \in G$.  
\end{definition}
Examples of groups include the permutations of a set and the general linear group $\textnormal{GL}(V)$ of invertible operators over a vector space $V$, both equipped with the usual composition and inversion of functions. A further example that will be relevant in this work is the \emph{unitary} group $\textnormal{U}(V) \subseteq \textnormal{GL}(V)$ associated to a Hilbert space $V$, consisting of operators $U$ satisfying $UU^{\dagger} = I$, where $I$ is the identity matrix.\footnote{Given a linear operator $U\colon H \rightarrow H'$ between Hilbert spaces, we denote by $U^\dagger \colon H' \rightarrow H$ its adjoint operator, defined by $\langle Ux , y \rangle = \langle x, U^\dagger y \rangle$ for all $x\in H$, $y \in H'$.} Groups satisfying $gh = hg$ for all $g,h \in G$ are deemed \emph{commutative}. The idea of a space $\mathcal{X}$ having $G$ as a group of symmetries is abstracted by the notion of group \emph{action}. 
\begin{definition}\label{actiondef}
An action by a group $G$ on a set $\mathcal{X}$ is a map $G \times \mathcal{X} \rightarrow \mathcal{X}$ denoted by $(g,x) \mapsto g \cdot x$, satisfying for all $g,h \in G, \ x \in \mathcal{X}$: 
\begin{center}
\begin{tabular}{ccc}
\emph{Associativity} & \hspace{1cm} & \emph{Identity} \\
$g\cdot (h \cdot x) = (gh) \cdot x$ & \hspace{1cm} &  $1 \cdot x = x$
 \end{tabular}
  \end{center}
A map $\varphi \colon \ \mathcal{X} \rightarrow \mathcal{Z}$ between sets acted upon by $G$ is called \emph{equivariant} if $\varphi(g \cdot x) = g \cdot \varphi(x)$ for all $g \in G, x\in \mathcal{X}$. It is called \emph{invariant} if moreover $G$ acts trivially on $\mathcal{Z}$ or, explicitly, if  $\varphi(g \cdot x) = \varphi(x)$. 
\end{definition}
In general, the following actions can be defined for arbitrary groups: $G$ acts on any set \emph{trivially} by $g \cdot x = x$, and $G$ acts on itself seen as a set via (left) \emph{multiplication} by  $g \cdot h = gh$. Further examples are $\textnormal{GL}(V)$ and $\textnormal{U}(V)$ acting on $V$ by evaluating operators. 

\subsection{Harmonic Analysis on Groups}

Harmonic analysis on groups \citep{folland2016course} generalizes standard harmonic analysis. We focus here on finite groups for simplicity, which are sufficient for practical applications. This avoids technicalities such as integrability conditions and continuity issues arising for infinite groups. We start by considering commutative groups and cover non-commutative ones in Section \ref{sec:nonabelian}.

Let $G$ be a finite commutative group of order $|G|$. Denote by $\free{G} = \mathbb{C}^G$ the free complex vector space generated by $G$. Intuitively, an element $x = (x_g)_{g \in G} \in \free{G}$ represents a complex-valued signal over $G$. The space $\free{G}$ is endowed with the \emph{convolution} product:
\begin{equation}\label{eq:convol}
(x \star y)_g = \sum_{h \in G}x_h y_{h^{-1}g},
\end{equation}
 and is acted upon by $G$ via $g \cdot x = \delta_g \star x = (x_{g^{-1}h})_{h \in G}$, where $\delta_g$ is the canonical basis vector. 
\begin{definition}
The \emph{dual} $G^\vee$ of $G$ is the set of homomorphisms $\rho: \ G \rightarrow \textnormal{U}(\mathbb{C})$, where $\textnormal{U}(\mathbb{C}) \subseteq \mathbb{C}$ is the group of unitary complex numbers equipped with multiplication. It is itself a group when equipped with pointwise composition $(\rho \mu)(g) = \rho(g) \mu(g)$.
\end{definition}
A homomorphism $\rho \in G^\vee$ intuitively represents a \emph{harmonic} over $G$, generalizing the familiar notion from signal processing. If we endow $\free{G}$ with the canonical scalar product $\langle x, y \rangle = \sum_{g \in G} \overline{x_g} y_g$, then $G^\vee \subseteq \free{G}$ forms an orthogonal basis with all the norms equal to $\sqrt{|G|}$. The linear base-change is, by definition, the Fourier transform over $\free{G}$: 
\begin{definition}\label{transform}
The \emph{Fourier transform} is the map $\free{G} \rightarrow \free{G^\vee}$, $x \mapsto \hat{x}$, defined for $\rho \in G^\vee$ as $\hat{x}_\rho = \langle \rho, x \rangle$.
\end{definition}
The Fourier transform is a linear isometry or, equivalently, a unitary operator, up to a multiplicative constant of $|G|$. Moreover, it exchanges the convolution product $\star$ over $\free{G}$ with the Hadamard product $\odot$ over $\free{G^\vee}$. Definition \ref{transform} generalizes the usual discrete Fourier transform in the following sense. For an integer $d>0$, consider the cyclc group $C_d$ with $d$ elements. Concretely, $C_d =\mathbb{Z}/d\mathbb{Z}$ is the group of integers modulo $d$ equipped with addition as composition. The dual $G^\vee$ consists of homomorphisms of the form $\mathbb{Z} / d\mathbb{Z} \ni g \mapsto e^{2 \pi \sqrt{-1} g k/d}$ for $k \in \{0, \cdots, d-1 \}$. Definition \ref{transform} specializes then to the familiar Fourier  transform.  

\subsection{Non-Commutative Harmonic Analysis}\label{sec:nonabelian}
So far, we have assumed that $G$ is commutative. In this section we briefly discuss the extension of Fourier theory to non-commutative groups. This however requires more elaborate theoretical tools, which we now introduce. To begin, in order to perform harmonic analysis on general groups it is necessary to discuss unitary representations. The latter will play the role of \textit{matrix-valued harmonics}. 
\begin{definition} 
A \emph{unitary representation} of $G$ is an action by $G$ on a finite-dimensional Hilbert space $V$ via unitary operators or, in other words, a homomorphism $\rho_V: G \rightarrow \textnormal{U}(V)$. A unitary representation is said to be \emph{irreducible} if $V$ does not contain any non-trivial\footnote{The trivial sub-representations of $V$ are $0$ and $V$.} sub-representations. 
\end{definition}
We denote by $\irr{G}$ the set of all irreducible representations of $G$ up to isomorphism. Moreover, for a vector space $V$ we denote by $\textnormal{End}(V)$ the space of its linear operators.   

\begin{definition}
The \emph{Fourier transform} is the map $\free{G} \rightarrow \bigoplus_{\rho_V \in \irr{G}} \textnormal{End}(V)$, $x \mapsto \hat{x}$, defined for $\rho_V \in \irr{G}$ as: 
\begin{equation}\label{fouriernc}
\hat{x}_{\rho_V} = \sum_{g \in G}
\rho_V(g)^{\dagger} x_g \in \textnormal{End}(V).  
\end{equation}
\end{definition}
This generalizes Definition \ref{transform} since for a commutative group, $\rho_V$ is irreducible if, and only if, $\textnormal{dim}(V) = 1$. Analogously to the commutative setting, the Fourier transform exchanges the convolution product $\star$ with the point-wise operator composition, which we still denote by $\odot$. Moreover, the Fourier transform is a unitary operator up to a multiplicative constant of $|G|$ with respect to the normalized Hilbert-Schmidt scalar product on $\textnormal{End}(V)$, given by $\langle A, B \rangle = \textnormal{dim}(V) \  \textnormal{tr}(A^\dagger B)$. The norm associated to the Hilbert-Schmidt scalar product is the Frobenius norm. The relations between irreducible unitary representations coming from the unitarity of the Fourier transform are known as \emph{Schur orthogonality} relations. 

\section{Theoretical Results}
\label{sec:theoretical-results}
We now present the primary theoretical contributions of this work. Concretely, we demonstrate that if certain parametric functions are invariant to a group then their weights must almost coincide with harmonics,  i.e. irreducible unitary group representations. We start by introducing general algebraic notions and principles, and then proceed to specialize them to machine learning scenarios. 

Let $G$ be a finite group, $\mathcal{H}$ be a set, and $V_1, \ldots, V_k$ be complex finite-dimensional Hilbert spaces. In what follows, we will consider the space:
\begin{equation}
\mathcal{W} = \free{G} \otimes \bigoplus_i \textnormal{End}(V_i) \simeq \bigoplus_i \textnormal{End}(V_i)^{\oplus G}.
\end{equation}
$\mathcal{W}$ is a Hilbert space when endowed with the scalar product given by the product of the canonical scalar product over $\free{G}$ and the normalized Hilbert-Schmidt scalar products over $\textnormal{End}(V_i)$. For $W \in \mathcal{W}$, we will denote each of its components as $W_i = (W_i(g))_g \in \textnormal{End}(V_i)^{\oplus G}$. Moreover, we will often interpret elements $W \in \mathcal{W}$ as linear maps $\free{G} \rightarrow \bigoplus_i \textnormal{End}(V_i)$ via $W(x) = \sum_{g \in G} W(g)x_g$ for $x \in \free{G}$, where $W(g) = (W_i(g))_i$. Note that $G$ acts on the left tensor factor of $\mathcal{W}$ while for every $i$, $\textnormal{U}(V_i)$ acts on the right tensor factor of $\free{G} \otimes \textnormal{End}(V_i)$ by composition of operators.     

\begin{example} 
Concretely, given coordinates on $V_i \simeq \mathbb{C}^{d_i}$ for $i=1, \ldots, k$, an element of $\mathcal{W}$ consists of a $k$-tuple of tensors $W_i \in \mathbb{C}^{d_i \times d_i \times |G|} \simeq \textnormal{End}(V_i)^{\oplus G}$. For example, the dihedral group $D_n$ of isometries of a regular $n$-gon has $2n$ elements. For $n$ odd, it has $2$ irreducible unitary representations of dimension $1$, and $(n-1)/2$ of dimension $2$, implying that $\mathcal{W}$ consists of a direct sum of copies of  $\mathbb{C}^{2n}$  and $\mathbb{C}^{2 \times 2 \times 2n}$.
\end{example}
\begin{definition}\label{def:symm}
We say that a map $\varphi \colon  \mathcal{W} \rightarrow \mathcal{H} $ has \emph{unitary symmetries} if for all $W,W' \in \mathcal{W}$ such hat $\| W_i \| = \|  W_i' \|$ for all $i$ and $\varphi(W) = \varphi(W')$, we have that for every $i$ there exists a unitary operator $U_i \in \textnormal{U}(V_i)$ such that $W_i = U_i \cdot W_i'$.    
\end{definition}

In the context of machine learning, $\mathcal{H}$ will represent the \emph{hypothesis space}, consisting of functions the model can learn. On the other hand, $\varphi$ will represent the parametrization of such hypotheses, with its domain $\mathcal{W}$ being the space of weights. Each component $\textnormal{End}(V_i)$ of $\mathcal{W}$ will be responsible for parametrizing a computational unit, i.e. a complex-valued \emph{neuron} in the language of neural networks. For commutative groups, we simply have $V_i = \mathbb{C} \simeq \textnormal{End}(V_i)$. In general, $\textnormal{End}(V_i)$ can be thought of as parametrizing matrix-valued signals, which, as mentioned in Section \ref{sec:overview}, are computed by neural units sometimes referred to as \emph{capsules} \citep{hinton2018matrix, sabour2017dynamic}. Lastly, the components of $\free{G}$ will represent the input space. The fact that the latter consists of scalar signals over $G$ is a simplification of several practical scenarios. However, the results of this section can be generalized, to an extent, to signals over a set acted upon by $G$ -- see Section \ref{rem:gract}. The following is an algebraic principle at the core of this work. 

\begin{theorem}\label{prop:mainnew}
Suppose that $\varphi \colon \mathcal{W} \rightarrow \mathcal{H}$ has unitary symmetries and that for some $W \in \mathcal{W}$ the following holds:
\begin{itemize}
  \setlength\itemsep{0.em}
\item $\varphi(g \cdot W ) = \varphi(W)$ for all $g \in G$. 
\item $W_i$, seen as a linear map $\free{G} \rightarrow \textnormal{End}(V_i)$, is surjective for all $i$.
\end{itemize}
Then for every $i$ there exist $W_i' \in \textnormal{End}(V_i)$ and an irreducible unitary representation $\rho_i: \ G \rightarrow \textnormal{U}(V_i)$ such that for all $g \in G$,
\begin{equation}
W_i(g) = W_i' \rho_i(g)^\dagger. 
\end{equation}
\end{theorem}

We refer to the Appendix for a proof. Note that the surjectivity assumption implies the constraint $\textnormal{dim}(V_i)^2 \leq |G|$ for all $i$. As a consequence of the result above, the full Fourier transform arises with an additional orthogonality assumption. 
\begin{corollary}\label{mainpropnc}
Suppose that $\varphi \colon \mathcal{W} \rightarrow \mathcal{H}$ has unitary symmetries and that for some $W \in \mathcal{W}$ the following holds:
\begin{itemize}
  \setlength\itemsep{0.em}
\item $\varphi(g \cdot W ) = \varphi(W)$ for all $g \in G$. 
\item $W$ is unitary up to a multiplicative constant, i.e. $W^\dagger W = |G| I $. 
\end{itemize}
Then $W$ is the Fourier transform up to composing each of the components $W_i$ by an operator with Frobenius norm equal to $1$.
\end{corollary} 
We refer to the Appendix for a proof. Again, the orthogonality assumption implies that $V_1, \ldots, V_k$ are the ambient Hilbert spaces of all the irreducible unitary representations of $G$ up to isomorphism, and in particular $\sum_i \textnormal{dim}(V_i)^2 = |G|$. 

We now wish to discuss the other crucial assumption of Theorem \ref{prop:mainnew} requiring that $\varphi(g \cdot W) = \varphi(W)$ for all $g \in G$, which is reminiscent of invariance. However, when $\mathcal{H}$ is a space of functions, we are typically interested in models that are invariant in the input variable rather than the weight variable. Therefore, we introduce the following condition, aimed at reconciling inputs and weights. To this end, suppose that $\mathcal{H}$ is a set of functions $\mathcal{X} \rightarrow \mathcal{Y}$, where $\mathcal{X}$ is a set acted upon by $G$ and $\mathcal{Y}$ is a set. We adhere to the notation $ \varphi(W, x) = \varphi(W)(x)$, $x \in \free{G}$, $W \in \mathcal{W}$, for simplicity.

\begin{definition}\label{defj:adj}
 We say that $\varphi$ satisfies the \emph{adjunction property} if $\varphi(W, g \cdot x) = \varphi(g^{-1} \cdot W, x)$ for all $x\in \mathcal{X}, g \in G$. 
\end{definition}
The adjunction property implies that if $\varphi(W, x)$ is invariant in $x$, then $\varphi(g \cdot W, x) = \varphi(W, x)$ for all $x, g$, recovering the assumption of Theorem \ref{prop:mainnew}.

\subsection{Extensions}\label{rem:gract}

As explained above, the tensor component $\free{G}$ of $\mathcal{W}$ typically represents the input space of a given machine learning model. However, it is often the case that data does not consist of signals over $G$.  This happens, for example, in the case of image data acted upon by the cyclic group via rotations (see Figure \ref{fig:feature-examples}), since the pixel plane is composed of several copies of $G$.

One can consider the more general scenario when data consist of complex signals over a finite set $\mathcal{S}$ acted upon by $G$, therefore replacing $\free{G} = \mathbb{C}^G$ by $\mathbb{C}^\mathcal{S}$. Assuming the action over $\mathcal{S}$ is \emph{free}, meaning that $g \cdot s = s$ implies $g=1$, $\mathcal{S}$ can be decomposed into copies of $G$ deemed \emph{orbits}. Specifically, there is an equivariant isomorphism $\mathcal{S} \simeq G \sqcup \cdots \sqcup G$, which in turn induces a linear isomorphism $\mathbb{C}^\mathcal{S} \simeq \free{G}^{\oplus p}$, where $p$ is the number of orbits. The results from this section can be extended to this scenario by applying all the arguments to the copies of $\free{G}$ separately, each of which will serve as a domain for its set of irreducible unitary representations.

When the action is not free, it is necessary to take into account \emph{stabilizers}, i.e. $g \in G$ such that $g \cdot s = s$ for some $s \in \mathcal{S}$. Roughly speaking, we expect that the results from this section can be adapted to an extent, obtaining unitary representations ``up to stabilizers". However, the precise meaning of the latter has yet to be clarified, and goes beyond the scope of this work. 

\subsection{Examples}\label{sec:exmp}
In this section, we provide examples of machine learning models with unitary symmetries. As anticipated, in the context of machine learning $\mathcal{H}$ and $\mathcal{W}$ represent the hypothesis space and the parameter space, respectively. Indeed, in what follows $\mathcal{H}$ will consist of functions of the form $\free{G} \rightarrow \mathcal{Y}$ for some codomain $\mathcal{Y}$, and we will adhere to the notation from Definition \ref{defj:adj} accordingly. All the models considered in this section satisfy the adjunction property.

\subsubsection{Spectral Networks}\label{sec:specnet}
We start by considering \emph{Spectral Networks} -- a class of polynomial machine learning models that inspired this work -- and to which our theory applies naturally. These models were introduced by \cite{sanborn2023bispectral} in cubic form and for commutative groups. Here, we generalize them to arbitrary degree and to the non-commutative setting. Spectral Networks are grounded in the algebraic invariant theory of $\free{G}$. We review the latter in the Appendix (Section \ref{sec:specinv}), including simple proofs for finite groups of classical results. The overall idea behind Spectral Networks is to approximate the $n$-order polynomial invariants -- deemed spectral invariants --  of $\free{G}$ for an \emph{unknown} group $G$. Specifically, suppose that $V_1, \ldots, V_k$ are the ambient Hilbert spaces of the irreducible unitary representations of $G$. Given a multi-index $\underline{i} = (i_1, \ldots, i_n) \in \{1, \ldots, k \}^n$, the Spectral Network of order $n$ is defined as the collection of parametric maps $\varphi_{\underline{i}}(W, \cdot):  \ \free{G}  \rightarrow \textnormal{End}(V_{i_1} \otimes \cdots \otimes V_{i_n})$: 
\begin{equation}\label{eq:ncspecnet}
\varphi_{\underline{i}}(W, x) =  W_{i_1} (x) \otimes \cdots \otimes W_{i_n}(x) \  \left( W_{i_1}^{\dagger} \odot   \cdots \odot W_{i_n}^{\dagger} \right) (\overline{x}),
\end{equation}
where $W=\oplus_iW_i \in \mathcal{W} = \free{G} \otimes  \oplus_i \textnormal{End}(V_i)$, $\odot$ denotes the $G$-wise tensor product of operators, and $\overline{x}$ denotes the component-wise conjugate of $x$. For a commutative $G$, since $V_i = \mathbb{C}$ for all $i$, the above expression reduces to $\varphi_{\underline{i}}(W, x) = W_{i_1}x  \cdots \ W_{i_n} x  \ \overline{W_{i_1} \odot \cdots \odot W_{i_n} \ x}$. For $n=1$ Spectral Networks are called Power-Spectral Networks, and for a commutative $G$ they take the form $\varphi_i(W, x) = | W_i x |^2$. The latter can be simply interpreted as a linear model followed by a function given by the squared absolute value. Even though the latter is uncommon in machine learning, it has appeared in models of biological neural networks \citep{adelson1985spatiotemporal}. 

For simplicity, and without loss of generality, we will consider only the Spectral Networks involving a single unitary representation; that is, we will focus on constant multi-indices $\underline{i} = (i, \ldots, i)$ in Equation \ref{eq:ncspecnet}. To this end, let $V$ be a finite-dimensional Hilbert space and $\mathcal{H}$ be the set of functions $\free{G} \rightarrow \textnormal{End}(V^{\otimes n})$ for some $n \in \mathbb{N}$. We also set $\mathcal{W} = \free{G} \otimes \textnormal{End}(V)$. The following is proved in the Appendix.

\begin{proposition}\label{prop:specunitmain}
Consider the Spectral Network given by $\varphi(W, x) = W(x)^{\otimes n} \ W^{\dagger \odot n}(\overline{x})$. Then $\varphi$ has unitary symmetries. 
\end{proposition}
 
\subsubsection{McCulloch-Pitts Neurons and Deep Networks}

While Spectral Networks provide the most direct application of our theory, in this section we discuss the most common and fundamental neural network primitives in deep learning: the fully-connected McCulloch-Pitts neuron \citep{mcculloch1943logical} and the deep neural network. We consider models with complex coefficients and focus on commutative groups, i.e. all the Hilbert spaces $V_i$ from Definition \ref{def:symm} will be equal to $\mathbb{C}$. A McCulloch-Pitts neuron has the form $\varphi(W,x) = \sigma(W x )$, where $\sigma \colon \mathbb{C} \rightarrow \mathcal{Y}$ is a map playing the role of an activation function and $W \in \mathcal{W} = \free{G}$ is the weight vector. For $\sigma(z) = |z|^2$, the McCulloch-Pitts neuron reduces to a commutative Power-Spectral Network, i.e. a Spectral Network with $n=1$.  The hypothesis space $\mathcal{H}$ consists of functions $\free{G} \rightarrow \mathcal{Y}$. 

\begin{proposition}\label{lemm:mcculloch}
Consider a map $\sigma \colon \mathbb{C} \rightarrow \mathcal{Y}$ and let $\varphi(W,x) = \sigma(W x )$. Suppose that $0 \in \mathbb{C}$ is isolated in its fiber of $\sigma$, i.e. there exists an open subspace $O \subseteq \mathbb{C}$ such that $\sigma^{-1}(\sigma(0)) \cap O = \{ 0 \}$. Then $\varphi$ has unitary symmetries.
\end{proposition}
We refer to the Appendix for a proof. The above assumption on $\sigma$ is satisfied by popular activations functions from neuroscience and machine learning, such as the sigmoid and the leaky Rectified Linear Unit (ReLU), applied after taking complex absolute value. Moreover, any non-constant holomorphic map $\sigma \colon \mathbb{C} \rightarrow \mathbb{C}$ satisfies the assumption, since its fibers are discrete by the Analytic Continuation Theorem. 

\begin{example} 
Consider the \emph{logistic regressor} with complex weights, corresponding to a McCulloch-Pitts neuron with a sigmoid activation function, i.e., $\sigma(z) = \frac{1}{1 + e^{-|z|}} + b$, where $b \in \mathbb{R}$ is the bias term. Since $\sigma^{-1}(\sigma(0)) = \{0\}$, Proposition \ref{lemm:mcculloch} applies. In particular, if the logistic regressor $\varphi(W , x) = \sigma(Wx)$ is invariant in $x$ to the cyclic group $C_d = \mathbb{Z} / d \mathbb{Z}$, from Theorem \ref{prop:mainnew} it follows that, if $W \not = 0$, then $W(g) = u \ e^{- 2 \pi \sqrt{-1}g/d}$ for all $g = 0, \dots, d-1$ and for some nonzero $u \in \mathbb{C}$. 
\end{example}

Next, we discuss the case of classical deep neural networks. We model the latter as $\varphi(W, x) = \chi(\sigma(W(x)))$, where $W \in \free{G} \otimes \mathbb{C}^k$ represents the weights of the first layer containing $k$ neurons, $\sigma\colon \mathbb{C} \rightarrow \mathbb{R}$ represents the activation function, computed coordinate-wise on $\mathbb{C}^k$, and $\chi \colon \mathbb{R}^k \rightarrow \mathbb{R}$ encompasses all the layers after the first one. Therefore, the first layer is assumed to be complex-valued, as required by our theory, while the subsequent layers are allowed to take real values, as typical in practice. Since the weights $W$ only account for the first layer, our main results will guarantee that invariant deep neural networks recover the Fourier transform in that layer, which is consistent with empirical observations \citep{olah2020overview}.   

Differently from Section \ref{sec:theoretical-results}, for the next result we will restrict $\mathcal{W}$ to the subspace of $\free{G} \otimes \mathbb{C}^k$ consisting of $W$ such that the components $W_i$ are orthonormal. This implies, in particular, the constraint $k \leq |G|$. Note that $\mathcal{W}$ is closed by the actions of $G$ and $\textnormal{U}(\mathbb{C})$. The orthonormality condition is anyway necessary in order to recover the full Fourier transform -- see Corollary \ref{mainpropnc}.  

\begin{proposition}\label{prop:deepnets}
Let $\mathcal{W} = \{ W \in \free{G} \otimes \mathbb{C}^k \ | \ \forall i,j \ \langle W_i, W_j\rangle = 0 \}$, consider maps $\sigma \colon \mathbb{C} \rightarrow \mathbb{R}$, $\chi \colon \mathbb{R}^k \rightarrow \mathbb{R}$, and let $\varphi(W,x) = \chi(\sigma(W(x)))$. Suppose that:
\begin{itemize}
\item There exists an open subspace $O \subseteq \mathbb{R}^k$ containing $\sigma(0)$ where $\chi$ is affine with distinct non-vanishing coefficients i.e., $\chi(z) = \sum_i a_i z_i + b$ for $z \in O$ with $0 \not = a_i \not = a_j$ for $i \not = j$.
\item $\sigma \in C^2(\mathbb{C})$ and $\partial_z \partial_{\overline{z}} \sigma  (0) \not = 0$.  
\end{itemize}
Then $\varphi$ has unitary symmetries. 
\end{proposition}  
We refer to the Appendix for a proof. The hypothesis on $\chi$ is motivated by the fact that typical (real-valued) deep networks have piece-wise linear activation functions, such as (leaky) ReLU. As such, they define piece-wise affine maps and therefore are affine when restricted to appropriate open subspaces. Moreover, the hypothesis on the coefficients $a_i$ in Proposition \ref{prop:deepnets} is \emph{generic}, meaning that it defines an open dense subset of $(a_1, \ldots, a_k) \in \mathbb{R}^k$. 

\subsection{Group Recovery}\label{sec:grrecov}
Corollary \ref{mainpropnc} allows us to recover the group structure of $G$ up to isomorphism from the weights of a map with unitary symmetries. In other words, this enables the recovery of an unknown group in a data-driven manner from the weights of an invariant machine learning model, addressing the problem of symmetry discovery discussed in Section \ref{sec:overview}. The procedure was originally suggested and validated empirically by \cite{sanborn2023bispectral}. To this end, assume that $\varphi_i$ satisfies the requirements of Corollary \ref{mainpropnc}. Moreover, we introduce the additional assumption that $W_i(1) = I \in \textnormal{End}(V_i)$ for all $i$. If that is the case, $W$ coincides exactly with the Fourier transform by Corollary \ref{mainpropnc}. This implies that the multiplication table of $G$ can be recovered by: 

\begin{equation}\label{recoveryalg}
gh = \underset{l\in G}{\textnormal{argmin}} \ \| W(g) \odot  W(h)  -   W (l)   \|,
\end{equation}
where  $\odot$ denotes the Hadamard product, i.e. component-wise operator composition. Note that this notation has a different meaning here than in Section \ref{sec:specnet}. Since the $W(g)$'s are orthogonal, the only possible values for the norms over which the minimum is performed are $0$ and $\sqrt{2 |G|}$.

The multiplication table of $G$ is a discrete object, while the weights $W \in \mathcal{W}$ can vary continuously. Therefore, it is natural to expect that the invariance condition ($g \cdot W = W$ for all $g \in G$) can be relaxed, while still recovering the multiplication table correctly. In what follows, we analyze relaxations of invariance and give bounds in which the group recovery algorithm holds. We start by introducing a quantity measuring how close a map is to having unitary symmetries. To this end, we assume that $\mathcal{H}$ is a metric space with distance function $\Delta: \mathcal{H} \times \mathcal{H} \rightarrow \mathbb{R}_{\geq 0}$. 
\begin{definition}
Given a map $\varphi \colon \mathcal{W} \rightarrow \mathcal{H} $, its \emph{unitarity defect} is defined for $\delta \in \mathbb{R}_{>0}$ as: 
\begin{equation}
\omega_\varphi(\delta) = \sup_{W, W'} \ \max_i  \ \inf_{U \in \textnormal{U}(V_i)} \ \  \| W_i - U \cdot W_i'  \|  , 
\end{equation}
where the supremum runs over all the $W, W' \in  \free{G} \otimes \textnormal{End}(V)$ of the same norm such that $\Delta(\varphi(W),  \varphi(W')) \leq \delta$.
\end{definition}
Note that $\varphi$ has unitary symmetries if, and only if, $\omega_\varphi(0) = 0$. We now state our main relaxation result. 
\begin{theorem}\label{prop:mainrelax}
Suppose that $\varphi \colon \mathcal{W} \rightarrow \mathcal{H}$ is a map and fix $W \in \mathcal{W}$. Denote $L = \| W^\dagger W - |G|I \|_\infty$, where $\| A \|_\infty = \max_{g,h}|A_{g,h}|$ is the uniform Frobenius norm for $G \times G$ matrices. Suppose that the following holds:
\begin{itemize}
  \setlength\itemsep{0.em}
 \item  For all $g\in G$, $\omega_{\varphi} \left( \Delta(\varphi(g \cdot W), \varphi(W)) \right) < \sqrt{\frac{1}{2} - \frac{L}{|G|}} \  \Big/ \left(\sqrt{|G| + L} + 1 \right)$. 
\item $L \leq \frac{|G|}{2}$. 
\item $W_i(1) = I \in \textnormal{End}(V_i)$ for all $i$. 
\end{itemize}
Then Equation \ref{recoveryalg} holds, i.e. the group recovery algorithm is correct.    
\end{theorem}
We refer to the Appendix for a proof. The assumption on $L$ in the above result is a relaxation of the unitarity assumption in Corollary \ref{mainpropnc} since $L=0$ if, and only if, $W$ is unitary up to a multiplicative constant. 

We provide an explicit bound for the unitarity defect of the McCulloch-Pitts neurons discussed in Section \ref{sec:exmp}. To this end, let $\mathcal{H}$ be the space of continuous functions defined on the unit sphere in $\free{G}$ equipped with the uniform metric (i.e., the $L_\infty$ distance) as $\Delta$. 
\begin{proposition}\label{prop:modulusculloch}
Let $W \in \free{G}$ and consider $\varphi(W,x) = \sigma(W x)$. Suppose that the activation function $\sigma \colon \mathbb{C} \to \mathbb{C}$ is continuous and satisfies the following coercivity condition: there exist constants $C \in \mathbb{R}_{>0}$, $n \in \mathbb{N}$ such that for every $x \in \mathbb{C}$, $|\sigma(0) - \sigma(x)| \geq C \ |x|^n$. Then the unitarity defect of $\varphi$  satisfies for $\delta < C$: 
\begin{equation}\label{eq:mccullochbound}
\omega_\varphi(\delta)  \leq \sqrt{2\left( 1- \sqrt{1 -  \left( \frac{\delta}{C} \right)^{\frac{2}{n}}} \right) }. 
\end{equation}
\end{proposition}
We refer to the Appendix for a proof. Note that the coercivity condition from above plays the role of the assumption on $\sigma$ from Proposition \ref{lemm:mcculloch}.

\subsubsection{Implementation}
\label{sec:experimental-results}
We now empirically explore the theory developed in this paper and demonstrate that Spectral Networks are able to recover the group structure in practice. We implement a non-commutative Power-Spectral Network $\varphi_i(W, x) = W_ix \ W^\dagger_i \overline{x}$ with weights $W \in \mathbb{C}^{d_i \times d_i \times d}$, where $d = |G|$ is the cardinality of the given group and $d_1, \ldots, d_k$ are the dimensions of its irreducible unitary representations. As discussed at the beginning of Section \ref{sec:grrecov}, we force $W_i(1) = I \in \mathbb{C}^{d_i \times d_i}$, where the index $1$ is arbitrarily chosen. 

\begin{figure}[t]
\centering
\subfigure[$C_6 \simeq C_2 \times C_3$]{\includegraphics[trim={0 2cm 0 0}, width=.3\linewidth]{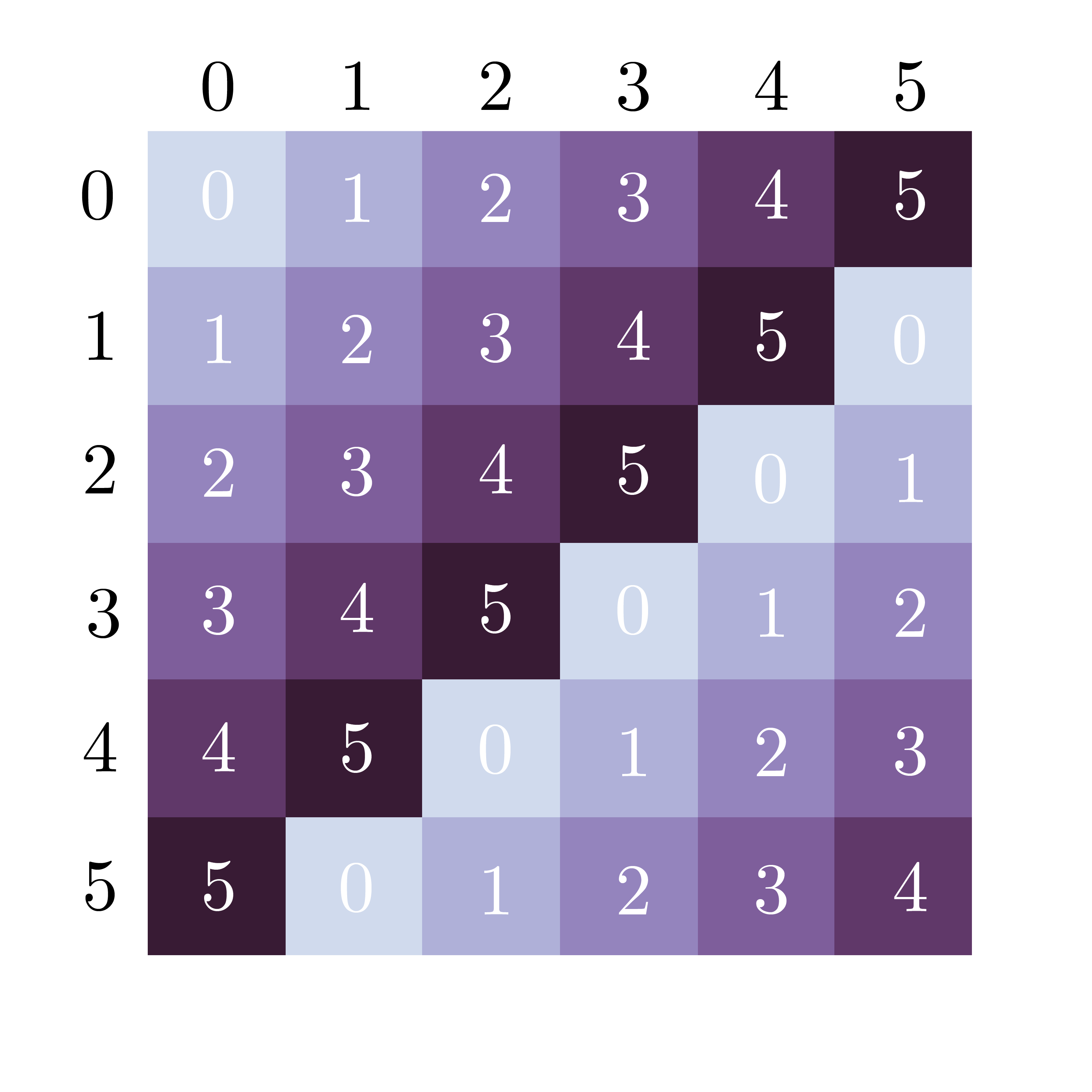}}      
\hspace{1em}
\subfigure[$C_2 \times C_2 \times C_2$]{\includegraphics[trim={0 2cm 0 0}, width=.3\linewidth]{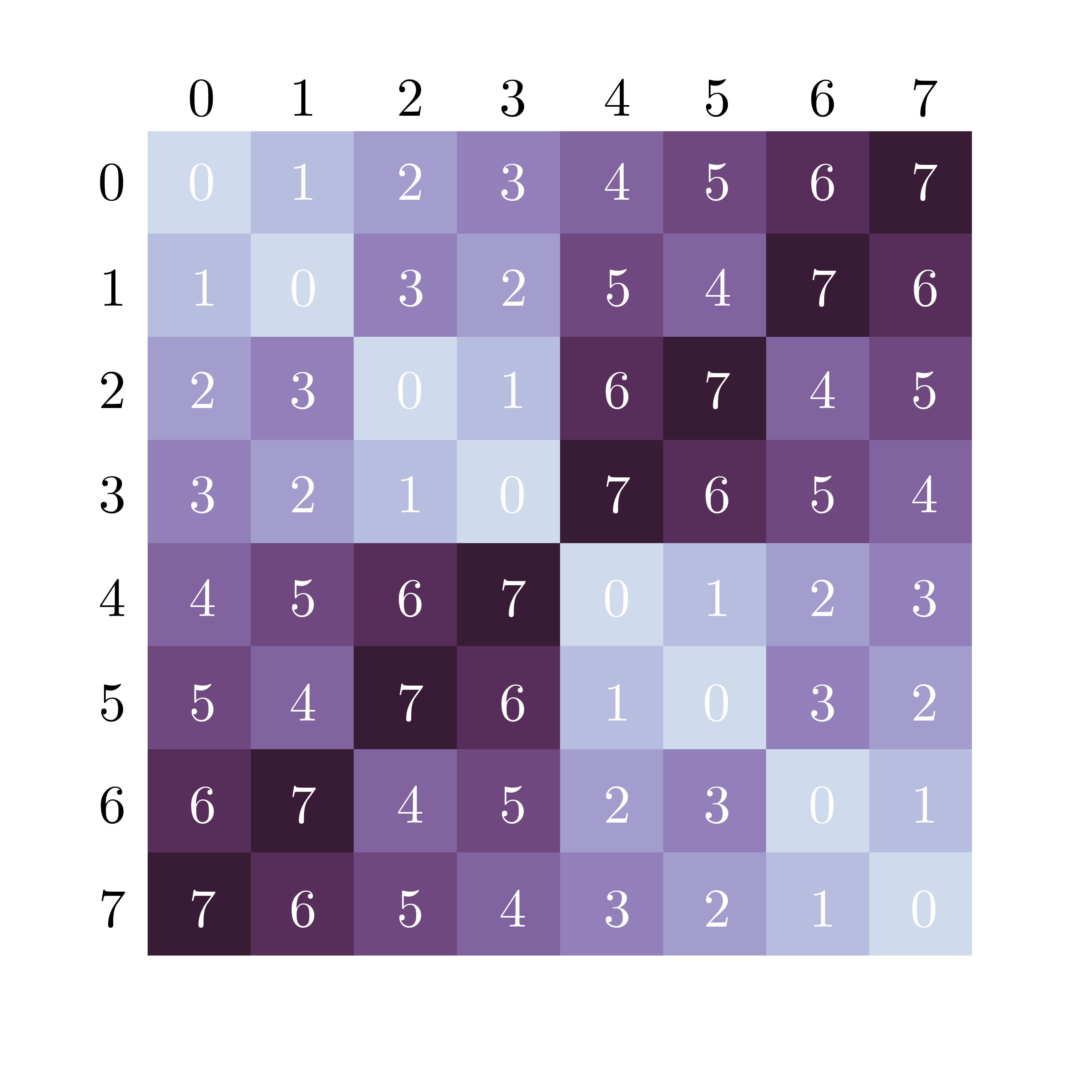}}
\hspace{1em}
\subfigure[$D_3 \simeq S_3$]{\includegraphics[trim={0 2cm 0 0}, width=.3\linewidth]{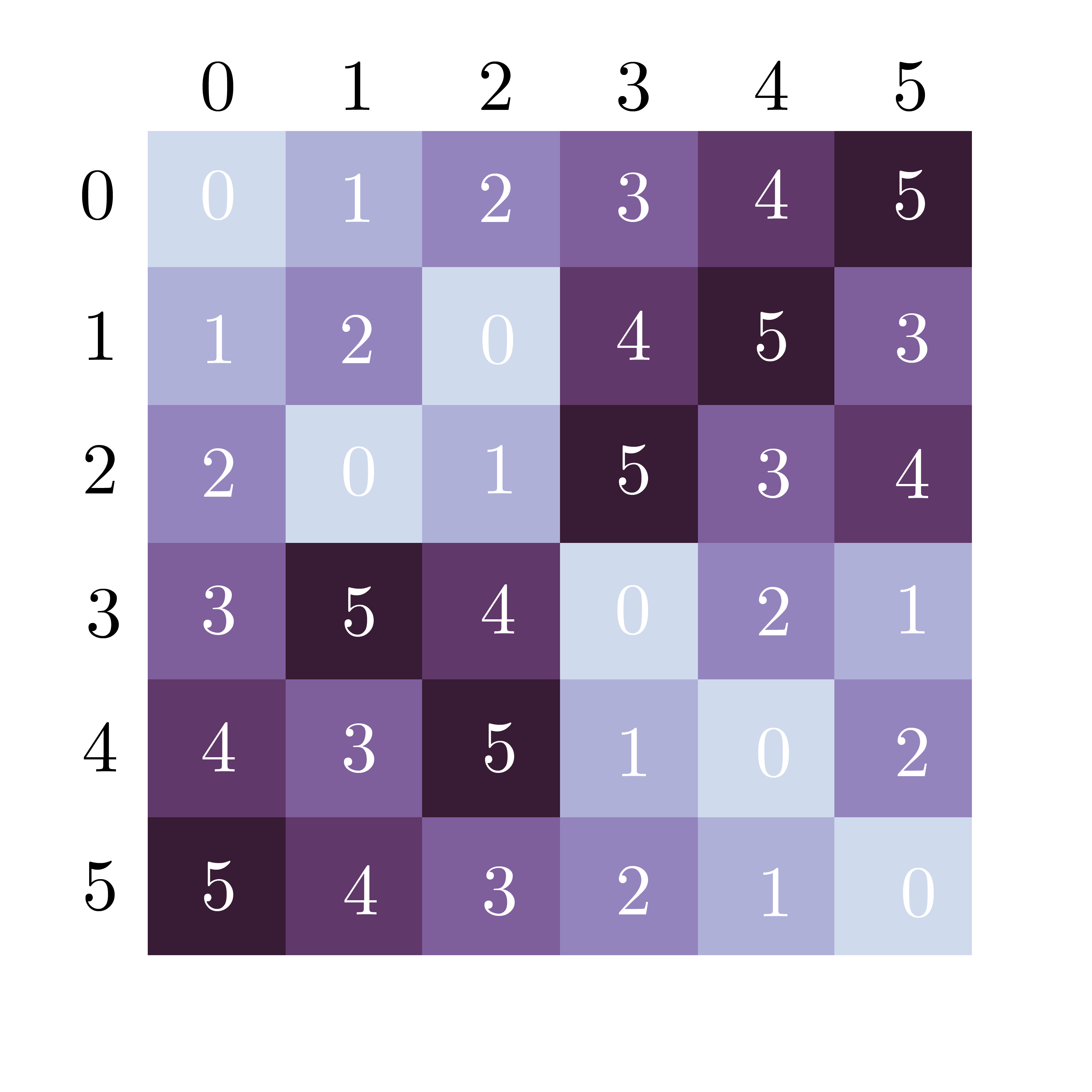}}
\caption{Multiplication tables inferred by Power-Spectral Networks for various groups. Group elements are labeled with integers. The table is symmetric if, and only if, the group is commutative.} 
\label{fig:cayley}
\end{figure}

Following \cite{sanborn2023bispectral}, we train the model via \emph{contrastive learning} \citep{jaiswal2020survey}. Namely, given a finite dataset $\mathcal{D}$ of pairs $(x,y)$, where $x,y \in \free{G} \simeq  \mathbb{C}^d$ and $x = g \cdot y$ for an unknown $g \in G$, the objective optimized by the model is:
\begin{equation}\label{eq:objcomm}
\mathcal{L}(W) = \sum_{(x,y) \in \mathcal{D}} \sum_{i} \ \|\varphi_{i}(W, x) - \varphi_{i}(W, y) \|^2 + \eta \ \|dI    -  W W^\dagger  \|^2 , 
\end{equation}
where $\eta > 0$ is a hyper-parameter and $\| \cdot \|$ is the Frobenius norm. The first term in Equation \ref{eq:objcomm} encourages invariance with respect to $G$ while the second one encourages $W$ to be unitary. The model is trained via the Adam optimizer \citep{kingma2014adam}, which interprets the complex weights as real tensors of doubled dimensionality. The dataset $\mathcal{D} \subseteq \free{G}\oplus \free{G} $ is generated procedurally by first sampling $x$ from a standard Gaussian distribution over $\free{G} \simeq \mathbb{R}^{2d}$, then sampling $g \in G$ uniformly, and finally producing the datapoint $(x,y=g \cdot x) \in \mathcal{D}$. We provide a Python implementation of both the model and the experiments at a public repository (see Section \ref{sec:overview}). The code is available in both the PyTorch \citep{NEURIPS2019_9015} and the JAX \citep{jax2018github} frameworks. 

Once trained, we evaluate the model by checking whether the multiplication table $M \in \{1, \ldots, d \}^{d\times d}$ obtained via the group recovery algorithm described by Equation \ref{recoveryalg} coincides with the one of $G$. Since there is no canonical ordering on $G$, the table is recovered up to a permutation $\pi$ of $\{1, \ldots, d \}$ acting as $(\pi \cdot M)_{i,j} = \pi(M_{\pi^{-1}(i), \pi^{-1}(j)})$. Therefore, we check whether $\pi \cdot M$ coincides with the table of $G$ for all the permutations $\pi$. Figure \ref{fig:cayley} reports the (correct) multiplication tables obtained at convergence for both commutative and non-commutative groups. Specifically, we consider the cyclic group $C_6$, the product of cyclic groups $C_2 \times C_2 \times C_2$, and the dihedral group $D_3$, which is isomorphic to the symmetric group $S_3$. 

\begin{figure}[t]
\centering

 \subfigure[$C_5$]{\includegraphics[width=.45\linewidth]{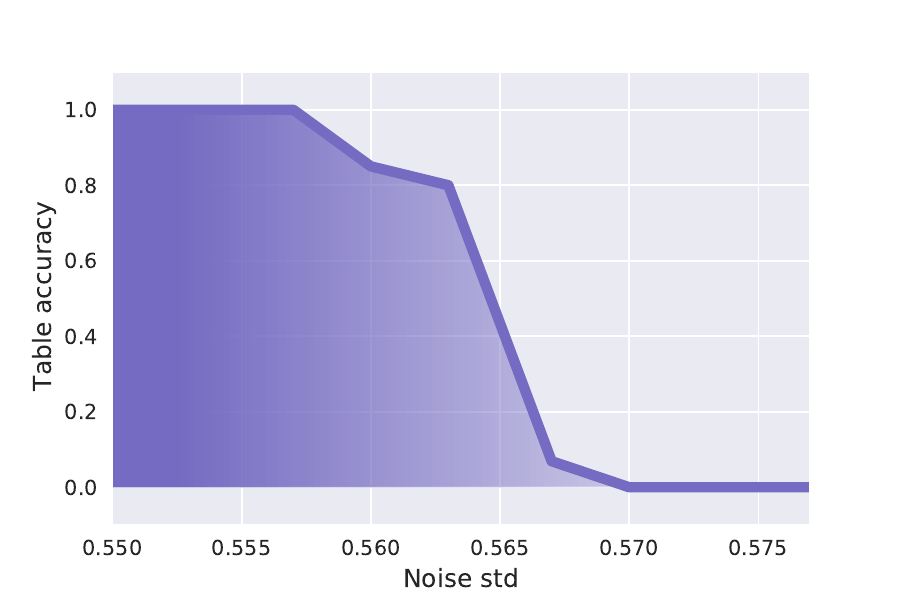} }
 \hspace{1em}
 \subfigure[$C_6$]{\includegraphics[width=.45\linewidth]{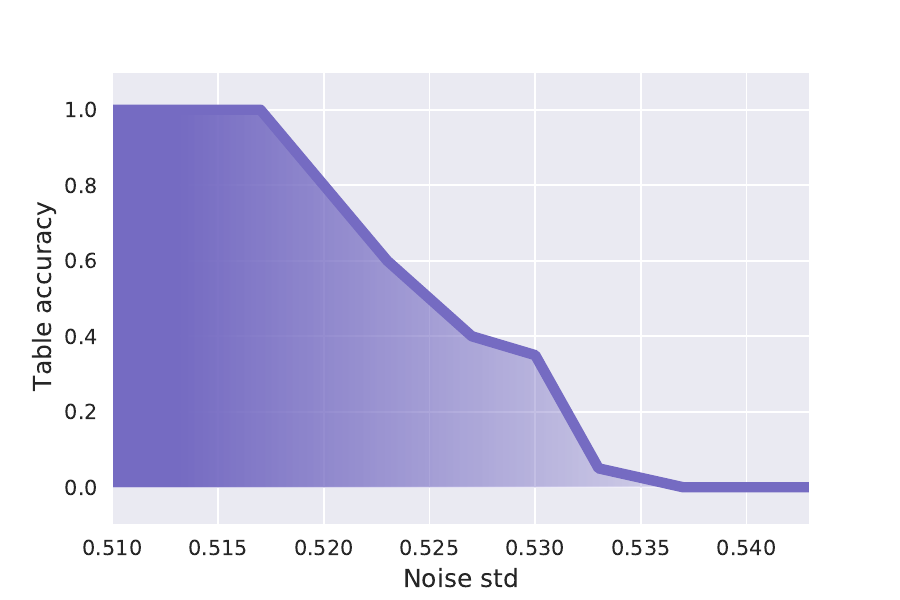}}
\caption{Accuracy for the recovery of the group multiplication tables across $20$ training runs as the amount of noise injected into data increases. Since the transition from $1.0$ to $0.0$ accuracy is sharp, we visualize only the noise regions where the values are non-trivial. } 
\label{fig:cayleybars}
\end{figure}

In order to validate empirically the robustness of the group recovery procedure, we additionally train the model on data corrupted by white noise, i.e. $\mathcal{D}$ consists of pairs $(x, y = g \cdot x + \epsilon)$, where $\epsilon$ is sampled from an isotropic Gaussian distribution. We vary the standard deviation of the latter and report in Figure \ref{fig:cayleybars} the number of times the multiplication table is recovered correctly across $20$ training runs -- a metric referred to as ``table accuracy". As can be seen, the group structure is recovered most of the times even with large amounts of noise -- up to $\sim 0.5$ standard deviation. The performance quickly degrades as the noise reaches a critical threshold. This shows empirically that the group recovery procedure is robust to noise, which is in line with the bounds from Theorem \ref{prop:mainrelax}.  

\section{Conclusions, Limitations, and Future Work}
\label{sec:discussion}

In this work, we proved that if a machine learning model of a certain kind is invariant to a finite group, then its weights are closely related to the Fourier transform on that group. We discussed how, as a consequence, the algebraic structure of an unknown group can be recovered from a model that is invariant. We established these results for both commutative and non-commutative groups, and discussed relaxed conditions under which the group recovery procedure holds. Our results represent a first step towards a mathematical explanation of universal features inferred by both biological and artificial neural networks. As such, this work contributes to the line of mathematical results stemming from biology \citep{sturmfels2005can}.

Due to its open-ended nature, this work is subject to a number of limitations and leaves directions open for future investigation. First,  our theory encompasses models with complex-valued weights, which are non-canonical in machine learning. Thus, exploring analogues of the theory over real numbers is an interesting direction that would fit more directly with current practices in the field. 

In addition, our theoretical framework encompasses learning models with unitary symmetries. The latter is a technical property satisfied by Spectral Networks and, to an extent, by traditional deep networks. However, it is not clear what other models fit into our framework, and whether the notion is general enough to accommodate other computational primitives fundamental to machine learning, such as attention mechanisms. This is an open question that is worthy of investigation.

Lastly, in this work we focused on groups and their associated harmonics. However, the representations within neural networks or biological systems often resemble imperfect, or \textit{localized}, versions of harmonics, i.e. wavelets, such as Gabors. Since wavelets do not describe group homomorphisms, our theory would need to be extended to accommodate this kind of locality. We suspect that this may be achieved by generalizing the framework to \emph{groupoids} -- an algebraic group-like structure that formalizes a locally-defined composition. This, however, goes beyond the scope of our work, and we leave it for future research. 

\acks
This work was supported by the Swedish Research Council, the Knut and Alice Wallenberg Foundation and the European Research Council (ERC-BIRD-884807).

\newpage 

\bibliography{biblio.bib}

\newpage 

\appendix

\section{Proofs of Theoretical Results}
\subsection{Proof of Theorem \ref{prop:mainnew}}
\begin{proof}
Since $\varphi$ has unitary symmetries and $\| g \cdot W_i \| = \| W_i\|$ for all $i$, it follows that for every $g \in G$ and every $i$ there exists $\rho_i(g) \in \textnormal{U}(V_i)$ such that:
\begin{equation}\label{eq:generalw}
g \cdot W_i =  \rho_i(g) \cdot W_i . 
\end{equation}
In particular, by considering the component with index $1 \in G$ on both sides of Equation \ref{eq:generalw}, we see that $W_i(g^{-1}) = W_i(1) \ \rho_i(g)$. We wish to show that $\rho_i$ is a homomorphism. To this end, for all $g,h \in G$ it holds that:
\begin{equation}\label{eq:generalproof}
 \rho_i(gh) \cdot W_i = (gh) \cdot W_i = g \cdot ( \rho_i(h) \cdot W_i ) = \rho_i(h) \cdot (g \cdot W_i)  = (\rho(g) \rho(h)) \cdot W_i.  
\end{equation}
By the surjectivity hypothesis, the set $\{W_i(g) \}_{g \in G}$ generates $\textnormal{End}(V_i)$ as a vector space. Therefore, Equation \ref{eq:generalproof} implies that $\rho_i(gh) = \rho_i(g) \rho_i(h)$, as desired. Lastly, note that $\rho_i$ is irreducible by the sujectivity assumption. Indeed, a non-trivial linear subspace of $V_i$ fixed by $\rho_i(g)$ for all $g$ would be sent by $W_i(g)$ into a fixed proper subspace due to Equation \ref{eq:generalw}. This contradicts the surjectivity of $W_i$. 
\end{proof}
\subsection{Proof of Corollary \ref{mainpropnc}}
\begin{proof}
Firstly, note that the unitarity assumption implies the surjectivity assumption from Theorem \ref{prop:mainnew}. Therefore, it follows that for every $i$, there exists an irreducible unitary representation $\rho_i \colon G \rightarrow \textnormal{U}(V_i)$ such that $W_i(g) = W_i(1) \ \rho_i(g)^\dagger$. We wish to show that if $i \not = j$ then $\rho_i$ and $\rho_j$ are non-isomorphic representations. If not, the orthogonality assumption implies that:
\begin{align}   
0  = \sum_{g \in G}  \overline{W_i(g)} \otimes W_j(g)  & = \sum_{g \in G}  \left( \overline{W_i(1)} \otimes W_j(1)  \right) \left(  \rho_i(g)^{\top} \otimes \rho_j(g)^\dagger    \right) = \\
& =  |G| \  \overline{W_i(1)} \otimes W_j(1),
\end{align}
where $\top$ denotes the transpose and where the last identity follows from the Schur orthogonality relations. But then $W_i(1)$ or $ W_j(1)$ is vanishing, which contraddicts the unitarity assumption. 

Lastly, in order to compute the Frobenius norm of $W_i(1)$, note that: 
\begin{equation}
|G| \ \| W_i(1) \|^2 = \sum_{g \in G} \| W_i(g) \ \rho_i(g) \|^2 = \sum_{g \in G} \| W_i(g) \|^2 = |G|,
\end{equation}
from which $\| W_i(1) \| = 1$ follows. 
\end{proof}

\subsection{Proof of Proposition \ref{prop:specunitmain}}
In order to prove this proposition, we will need some technical facts from matrix algebra. We start by showing the following uniqueness result. 

\begin{lemma}\label{polarlemm}
Let $d' \geq d$ and $A,B \in \mathbb C^{d' \times d}$. If $AA^{\dagger} = BB^{\dagger}$, then there exists a unitary matrix $U \in \mathbb{C}^{d \times d}$ such that $A = BU$. 
\end{lemma}

\begin{proof}
From the polar decomposition for matrices \cite[Theorem 3.1.9]{horn1994topics}, we know that:
\begin{equation} 
A = PV, \ B = Q W,
\end{equation}
where $P,Q \in \mathbb{C}^{d' \times d'}$ are Hermitian positive semidefinite and $V,W \in \mathbb{C}^{d' \times d}$ have orthonormal rows.  Also, $P^2 = AA^{\dagger} = BB^{\dagger} = Q^2$ from which it follows that $P = Q$ by uniqueness of square roots of positive semidefinite Hermitian matrices \cite[Theorem 7.2.6]{horn2012matrix}.  In particular, we have:
\begin{equation}
A = PV = QV = QWW^{\dagger}V = BU,
\end{equation}
with $U = W^{\dagger}V$ unitary. 
\end{proof}

\begin{remark}
We note that if $A$ and $B$ are real matrices, then the conclusion holds with $U$ being a real orthogonal matrix.
\end{remark}

Next, we show that positive semidefinite Hermitian matrices possess unique tensor roots.

\begin{lemma}\label{tensorroot}
Let $A, B \in \mathbb{C}^{d\times d}$ be Hermitian and positive semidefinite. If $A^{\otimes n} = B^{\otimes n}$
for some $n > 0$, then $A =B$. 
\end{lemma}

\begin{proof}
From the Spectral Theorem we know that:
\begin{equation}
A = U D U^{\dagger},
\end{equation}
where $U$ is unitary and $D$ is diagonal. Since $A^{\otimes n} = B^{\otimes n}$, it follows that $D^{\otimes n} = C^{\otimes n}$, where $C = U^{\dagger}BU$. Note that the (point-wise) Hadamard product of matrices is a submatrix of the tensor (Kronecker) product. In particular, the off-diagonal entries of $C$ must vanish. On the diagonal we have $(D_{i,i})^n = (C_{i,i})^n$ for every $i$, and therefore $D_{i,i} = C_{i,i}$ since they are non-negative. This shows that $C = D$, implying that $B = UDU^{\dagger} = A$.  
\end{proof}

Putting together the above lemmas, we obtain the following. 

\begin{lemma}\label{lemmnc}
Let $A_1, \ldots, A_k, B_1, \ldots, B_k \in \mathbb{C}^{d \times d}$. Suppose that for some $n>0$, for all $x \in \mathbb{C}^k$:  
\begin{equation}
\left( \sum_i x_i A_i  \right)^{\otimes n} \  \left( 
 \sum_i \overline{x}_i  A_i^{\dagger \otimes n}   \right) = \left( \sum_i x_i B_i  \right)^{\otimes n} \  \left( 
 \sum_i \overline{x}_i  B_i^{\dagger \otimes n}   \right).
\end{equation}
Then there exists a unitary matrix $U \in \mathbb{C}^{d \times d}$ such that $A_i = B_i U  $ for every $i$. 
\end{lemma}
\begin{proof}
By multilinearity of the tensor product we see that for all $x \in \mathbb{C}^{k}$: 
\begin{align}
&\sum_{i_1, \ldots, i_{n+1}}x_{i_1} \cdots x_{i_n} \overline{x}_{i_{n+1}} (A_{i_1} \otimes \cdots \otimes A_{i_n}) A_{i_{n+1}}^{\dagger \otimes n} = \\
= &\sum_{i_1, \ldots, i_{n+1}}x_{i_1} \cdots x_{i_n} \overline{x}_{i_{n+1}} (A_{i_1} A_{i_{n+1}}^{\dagger} ) \otimes \cdots \otimes ( A_{i_n} A_{i_{n+1}}^{\dagger} ) =   \\
= &\sum_{i_1, \ldots, i_{n+1}}x_{i_1} \cdots x_{1_n} \overline{x}_{i_{n+1}} (B_{i_1} B_{i_{n+1}}^{\dagger} ) \otimes \cdots \otimes ( B_{i_n} B_{i_{n+1}}^{\dagger} ).
\end{align}
Since a polynomial vanishes as function if and only if it is the zero polynomial, it follows that $(A_{i_1} A_{i_{n+1}}^{\dagger} ) \otimes \cdots \otimes ( A_{i_n} A_{i_{n+1}}^{\dagger} ) = (B_{i_1} B_{i_{n+1}}^{\dagger} ) \otimes \cdots \otimes ( B_{i_n} B_{i_{n+1}}^{\dagger} )$ for all $i_1, \ldots, i_{n+1}$, and in particular $(A_{i} A_{j}^{\dagger})^{\otimes n} = (B_{i} B_{j}^{\dagger})^{\otimes n}$ for all $i,j$. From Lemma \ref{tensorroot} we conclude that $A_iA_j = B_iB_j$ for all $i,j$, which can be rephrased as $A A^{\dagger} = B B^{\dagger}$, where $A, B$ are the $(d k) \times d $ matrices obtained by concatenating the $A_i$'s and $B_i$'s, respectively. From Lemma \ref{polarlemm} we conclude that $A = BU$ for a unitary $d\times d$ matrix $U$, as desired. 
\end{proof}

Proposition \ref{prop:specunitmain} now follows immediately from Lemma \ref{lemmnc} by setting $A_i = W(g_i)$ and $B_i = W'(g_i)$ for $g_i \in G$ and $W, W' \in \free{G} \otimes \textnormal{End}(V)$ of the same norm. 

\subsection{Proof of Proposition \ref{lemm:mcculloch}}
\begin{proof} Pick $W,W' \in \free{G}$ of the same norm such that $\varphi(W,x) = \varphi(W',x)$ for all $x \in \free{G}$. Given the open set $O \subseteq \mathbb{C}$ from the hypothesis on $\sigma$, consider $O' = \{ x \in \free{G} \ | \ Wx, W'x \in O \}$, which is open and non-empty since $0 \in O'$. For $x \in O'$, $Wx = 0$ implies $\varphi(W,x) =  \varphi(W', x) = 0$, from which it follows that $W'x = 0$ by definition of $O$. Therefore, $W$ and $W'$ share the same orthogonal complement, implying that $W' = \rho W$ for some $\rho \in \mathbb{C} $. Since $W$ and $W'$ have the same norm, we conclude that $\rho \in \textnormal{U}(\mathbb{C})$. 
\end{proof}

\subsection{Proof of Proposition \ref{prop:deepnets}}

\begin{proof}
Pick $W, W' \in \mathcal{W}$ such that $\| W_i\| = \| W_i' \|$ for all $i$ and $\varphi(W,x) = \varphi(W', x)$ for all $x \in \free{G}$. Given the open set $O \subseteq \mathbb{R}^k$ from the hypothesis on $\chi$, consider the set $O' = \{ x \in \free{G} \ | \ \sigma(W(x)), \sigma(W'(x)) \in O \}$, which is open and non-empty since $0 \in O'$. For $x \in O'$, an immediate calculation shows that:
\begin{equation}
\partial_x \partial_{\overline{x}} \ \varphi(W,x) = \sum_i a_i \  \partial_z \partial_{\overline{z}} \sigma(W_i \  x) \ W_i \otimes \overline{W_i},
\end{equation}
with the $a_i$'s being distinct, and similarly for $\varphi(W', x)$. By specializing to $x =0$, we deduce that:

\begin{equation}\label{eq:tensordense}
\bcancel{\partial_z \partial_{\overline{z}} \sigma(0)} \ \sum_i a_i \  W_i \otimes \overline{W_i} = \bcancel{\partial_z \partial_{\overline{z}} \sigma(0)} \ 
 \sum_i a_i  \ W_i' \otimes \overline{W_i'}. 
\end{equation}
Since both the sets $\{ W_i\}_i$ and $\{ W_i'\}_i$ are orthonormal by hypothesis on $\mathcal{W}$, both sides of Equation \ref{eq:tensordense} define a spectral decomposition, i.e. a decomposition into projections over orthonormal vectors. From the hypothesis on $\chi$ it follows that the eigenvalues $a_i \  \partial_z \partial_{\overline{z}} \sigma(W_i \  x)$ of the operators in Equation \ref{eq:tensordense} are distinct and non-vanishing, implying that the eigenvectors coincide up to a multiplicative scalar, i.e. $W_i = \rho_i W_i'$ for some $\rho_i \in \textnormal{U}(\mathbb{C})$, as desired. 
\end{proof}

\subsection{Proof of Theorem \ref{prop:mainrelax}}
\begin{proof}
Firstly, the definition of $L$ implies the inequalities $|G| - L \leq \| W(g)\|^2 \leq |G| + L$ and $|\langle W(g), W(h)\rangle| \leq L$ for all $g,h \in G$. In particular, we have: 
\begin{equation}
\|W(g) - W(h )\|^2 = \| W(g)\|^2 + \| W(h) \|^2 -2 \textnormal{Re}(\langle W(g), W(g) \rangle) \geq 2 |G| - 4L. 
\end{equation}
By hypothesis, for every $i$ and $g \in G$ there exists $\rho_i(g) \in \textrm{U}(V_i)$ such that: 
\begin{equation}\label{ineqc}
\| g^{-1} \cdot W_i - \rho_i(g) \cdot W_i \| < \frac{\sqrt{\frac{1}{2} - \frac{L}{|G|}}}{\sqrt{|G| + L} + 1}. 
\end{equation}
In particular, $\|  W_{i}(gh) - \rho_i(g)W_{i}(h)   \|$ is bounded by the same quantity for all $g,h \in G$. Therefore, via the triangle inequality we see that:
\begin{align}\label{ineqfinal}
& \| W_{i}(g)W_{i}(h) - W_{i}(gh) \| \leq \\
\leq & \|W_{i}(h) W_{i}(g) - \rho_i(g) W_{i}(h)\| + \| W_{i}(gh) - \rho_i(g)W_{i}(h) \| = \\
= & \underbrace{\|W_{i}(h)\|}_{\leq \sqrt{|G| + L}} \ \|W_{i}(g) - \rho_i(g)W_{i}(1) \|  + \| W_{i}(gh) - \rho_i(g)W_{i}(h) \| < \\ 
< & \sqrt{\frac{1}{2} - \frac{L}{|G|}}.
\end{align}
Equation \ref{ineqfinal} implies that: 
\begin{equation}
\| W(g) \odot W(h) - W(gh)  \| < \sqrt{|G|}\sqrt{\frac{1}{2} - \frac{L}{|G|}} = \frac{\sqrt{2|G| - 4L}}{2}.
\end{equation}
Since $\|  W(p)  - W(q)  \| \geq \sqrt{2|G| -4L}$ for all $p \not = q \in G$, $W(g) \odot W(h)$ is closer to $W(gh)$ than to any other $W(q)$ for $q \in G$, which immediately implies the desired result. 
\end{proof}

\subsection{Proof of Proposition \ref{prop:modulusculloch}}

In order to prove this proposition, we will need the following technical fact from linear algebra.

\begin{lemma}\label{lemmrelax}
Let $H$ be a finite-dimensional complex Hilbert space, $v,w \in H$ normal  and $\varepsilon \in \mathbb{R}$ such that $0 < \varepsilon < 1$. Suppose that for every normal $x$ orthogonal to $w$, it holds that $| \langle x, v\rangle | \leq \varepsilon$. Then there exists $\rho \in \textnormal{U}(\mathbb{C})$ such that: 
\begin{equation}\label{eq:sqrtsqrt}
\| v - \rho w \| \leq \sqrt{2\left( 1- \sqrt{1 -   \varepsilon^2 } \right) }. 
\end{equation}
\end{lemma}

\begin{proof}
Consider an orthogonal decomposition $v = \langle w_1, v \rangle  w_1 + \langle w_2,  v \rangle w_2$, where $w_1 \in w^\perp$ is normal and $w_2 = \rho w$ for some $\rho \in \textnormal{U}(\mathbb{C})$ such that $\langle w_2, v \rangle \in \mathbb{R}_{\geq 0}$. It follows that:
\begin{equation}
1 = \| v \|^2 = | \langle w_1, v \rangle |^2 +    \langle w_2, v \rangle ^2.
\end{equation}
The hypothesis implies then that $  \ \langle w_2, v \rangle\   \geq \sqrt{1 -   \varepsilon^2 } $. Therefore, we have: 
\begin{equation}
\| v - w_2 \|^2 = 2 - 2\langle w_2, v \rangle  \leq    2\left( 1- \sqrt{1 -   \varepsilon^2 } \right),
\end{equation}
as desired. 
\end{proof}

\begin{remark}
Note that the right-hand side of Equation \ref{eq:sqrtsqrt} is bounded by the concise quantity $\sqrt{2 \varepsilon}$.  
\end{remark}

We are now ready to prove Proposition \ref{prop:modulusculloch}.

\begin{proof}
Consider $\delta \in \mathbb{R}_{>0}$ and $W, W' \in  \free{G}$ of the same norm such that $\Delta(\varphi(W),  \varphi(W')) \leq  \delta$. The latter and the hypotheses together imply that if $x \in \free{G}$ is normal such that $W'  x = 0$, then:
\begin{equation}
C \ \left| W x    \right|^n \leq |  \sigma(0) - \sigma (W x )  | \leq \delta. 
\end{equation}
By Lemma \ref{lemmrelax}, there exists $\rho \in \textrm{U}(\mathbb{C}) $ such that: 
\begin{equation}
\| W - \rho W' \| \leq \sqrt{2\left( 1- \sqrt{1 -  \left( \frac{\delta}{C} \right)^{\frac{2}{n}}} \right) },
\end{equation}
from which the claim follows. 
\end{proof}

\section{Spectral Invariants}\label{sec:specinv}
In this section, we overview the theory of \emph{invariants} over $\free{G}$, i.e. (polynomial) maps $\free{G} \rightarrow \mathbb{C}$ that are invariant with respect to the action by $G$. To this end, we recall the following notion for a commutative group $G$. 

\begin{definition}\label{specdef}
Fix $n>0$ and $\underline{\rho} = (\rho_1, \ldots, \rho_{n}) \in (G^\vee)^{n}$. The \emph{spectrum of order} $n$ associated to $\underline{\rho}$ is defined for $x \in \free{G}$ as: 
\begin{equation}\label{eq:spectralformula}
\beta_{\underline{\rho}}(x) =  \hat{x}_{\rho_1} \cdots \hat{x}_{\rho_n}   \overline{\hat{x}}_{\rho_1 \cdots \rho_n}.
\end{equation}
\end{definition}
The spectra of order $n$ are invariant polynomials of degree $n+1$ containing one conjugate variable. The presence of the latter is necessary for invariance. For $n=1,2$ they are alternatively referred to as \emph{power spectra} and \emph{bispectra} respectively. Note that the power spectra reduce simply to $\beta_{\rho}(x) = |  \hat{x}_{\rho}  |^2 $, $\rho \in G^\vee$, and constitute a standard tool in signal processing. Bispectra, together with higher-order spectra, were first introduced by \cite{kakarala2012bispectrum}. It is immediate to see that spectra generate all the polynomial invariants of $\free{G}$  -- see also \cite[Theorem 2.1.4]{sturmfels2008algorithms}.
\begin{proposition}\label{prop:specgen}
The space of polynomial invariants of degree $n+1$ over $\free{G}$ (with one conjugate variable) is generated as a complex vector space by the spectra of order $n$.
\end{proposition}
\begin{proof}
 This follows from interpreting the invariance condition via the Fourier transform. Namely, given $\underline{\rho} \in (G^\vee)^{n+1}$ consider the monomial over $\free{G^\vee}$ defined by $\hat{x}_{\rho_1} \cdots\hat{x}_{\rho_n} \overline{\hat{x}}_{\rho_{n+1}}$. Since $g \cdot \hat{x} = (\overline{\rho(g)} \ \hat{x}_\lambda)_{\lambda \in G^\vee}$, the monomial is invariant if and only if $ \rho_1(g) \cdots \rho_n(g) \overline{\rho}_{n+1}(g)= 1$ for all $g \in G$, i.e. $\rho_{n+1} = \rho_1 \cdots \rho_n$. Since monomials linearly generate polynomials, the claim follows. 
\end{proof}

A remarkable aspect of spectra of even order is the fact that they jointly determine real (generic) elements of $\free{G}$ up to the action by $G$ -- a property known as \emph{completeness}. This was first shown by \cite{smach2008generalized}. For convenience, we report below a simple proof for finite commutative groups.  

\begin{proposition}\label{complete}
Fix $n$ even. Suppose that $x,y \in \mathbb{R}^G \subseteq \free{G}$ are such that $\hat{x}_\rho, \hat{y}_\rho \not = 0$ for all $\rho \in G^\vee$. If $\beta_{\underline{\rho}}(x) = \beta_{\underline{\rho}}(y)$ for all $\underline{\rho} \in (G^\vee)^{n}$ then $x = g \cdot y$ for some $g \in G$. 
\end{proposition}

\begin{proof}
By setting $\underline{\rho} = (1, \ldots ,1)$ we see that $\beta_{\underline{\rho}}(x) = \hat{x}_1^{n+1} = \beta_{\underline{\rho}}(y) =   \hat{y}_1^{n+1}\in \mathbb{R} \setminus \{0 \}$ and therefore $  \hat{x}_1 =  \hat{y}_1$ since $n$ is even. For $\rho \in G^\vee$, by setting $\underline{\rho} = (\rho, \overline{\rho}, 1, \ldots, 1)$, we see that $ \hat{x}_\rho \ \hat{x}_{\overline{\rho}} \  \hat{x}_1^{n-1} = | \hat{x}_\rho |^2  \ \hat{x}_1^{n-1}  =  | \hat{y}_\rho |^2 \ \hat{y}_1^{n-1} $ and therefore $| \hat{x}_\rho | = | \hat{y}_\rho |$. Note that here we relied on the fact that $x$ and $y$ are real. This implies that the following map $\eta: \ G^\vee \rightarrow \mathbb{C}$ takes values in $\textnormal{U}(\mathbb{C}) $: 
\begin{equation}
\eta(\rho) = \frac{ \hat{x}_\rho }{ \hat{y}_\rho }. 
\end{equation}
Now, $\eta(1) = 1$ since $ \hat{x}_1 =  \hat{y}_1 $. By setting $\underline{\rho} = (\rho, \mu, \overline{\rho \mu}, 1, \ldots, 1)$ we see that $\eta(\rho) \eta(\mu) = \eta(\overline{\rho \mu})$ for all $\rho, \mu \in G^\vee$ and therefore $\eta \in (G^\vee)^\vee$. Since the Fourier transform sends $G^\vee \subseteq \free{G}$ to $(G^\vee)^\vee \subseteq \free{G^\vee}$, there exists $g\in G$ such that $\eta(\rho) = \overline{\rho(g)}$. This means that $ \hat{x}_\rho  =  \overline{\rho(g)} \  \hat{y}_\rho $, which implies $x = g \cdot y$ by the equivariance properties of the Fourier transform.  
\end{proof}

Spectral invariants can be defined in the non-commutative case, but give rise to subtleties. First, we replace the group structure of $G^\vee$ with the tensor product $\otimes$ of unitary representations. However, $\irr{G}$ is not closed with respect to $\otimes$. This is circumvented by considering \emph{Clebsch-Gordan coefficients}, i.e. irreducible unitary sub-representations of tensor products. This leads to the following definition of operator-valued spectra. 

\begin{definition}\label{specdefnc}
Fix $n>0$ and $\underline{\rho} = (\rho_{V_1}, \ldots, \rho_{V_n}) \in \irr{G}^{n}$. The \emph{spectrum of order} $n$ associated to $\underline{\rho}$ is defined for $x \in \free{G}$ as: 
\begin{equation}\label{spectralformula}
\beta_{\underline{\rho}}(x) =  \hat{x}_{\rho_{V_1}} \otimes \cdots \otimes \hat{x}_{\rho_{V_n}} \left(  \hat{x}_{\rho_{T_1}}^{\dagger} \oplus \cdots \oplus \hat{x}_{\rho_{T_k}}^{\dagger} \right) \ \in \textnormal{End}(V_1 \otimes \cdots \otimes V_n),
\end{equation}
where the direct sum runs over the $k$ irreducible unitary representations appearing in an orthogonal decomposition $V_1 \otimes \cdots \otimes V_n = T_1 \oplus \cdots \oplus T_k$. 
\end{definition}
The completeness of spectra of order $n \geq 2$ (Proposition \ref{complete}) extends to the non-commutative case \citep{kakarala2009completeness}. The proof we provide of Proposition \ref{complete} translates almost identically to non-commutative groups, assuming $x,y \in \mathbb{R}^G$ have non-singular Fourier coefficients, i.e., $\hat{x}_{\rho_V}, \hat{y}_{\rho_V} \in \textnormal{GL}(V) \subseteq \textnormal{End}(V)$. The only substantial difference in the proof is that, in order to conclude that $\eta(\rho) = \rho(g)^\dagger $  for some $g \in G$, it is necessary to invoke Tannaka-Krein duality.

\end{document}